\documentclass[a4paper,11pt]{article}
\usepackage[utf8]{inputenc} 
\usepackage[T1]{fontenc}    
\usepackage{hyperref}       
\usepackage{url}            
\usepackage{booktabs}       
\usepackage{amsfonts}       
\usepackage{nicefrac}       
\usepackage{microtype}      
\usepackage{xcolor}         
\usepackage[square,numbers]{natbib}
\usepackage{verbatim}
\usepackage{algorithm}
\usepackage{dsfont}
\usepackage[noend]{algpseudocode}
\usepackage{amsmath}
\usepackage{amssymb}
\usepackage{amsthm}
\usepackage{hyperref}
\usepackage{graphicx}
\usepackage[title]{appendix}
\usepackage{multirow}
\usepackage{float}
\usepackage{authblk}

\newcommand{\mycomment}[1]{}
 
\newtheorem{thm}{Theorem}
\newtheorem{lem}[thm]{Lemma}

\newtheorem{cor}{Corollary}

\newcommand*{\CU}[1][\value{footnote}]{\footnotemark[2]}
\usepackage[left=1in,right=1in,top=1in,bottom=1in]{geometry}
\usepackage{bm}
\usepackage{scalerel}[2016/12/29]
\newcommand\sla{\scaleobj{0.8}{\leftarrow}}

\begin{document}
\title{Scores as Actions: a framework of fine-tuning diffusion models by 
continuous-time reinforcement learning\thanks{Working Paper}}

\author{Hanyang Zhao\thanks{Department of IEOR, Columbia University, New York, USA, \texttt{\{hz2684,hc3136,yao,wt2319\}@columbia.edu}}\ \ \  
Haoxian Chen\CU\ \ \ 
Ji Zhang\thanks{Department of CS, Stony Brook University, New York, USA, \texttt{ji.zhang@stonybrook.edu}}\ \ \
David D. Yao\CU \ \ \
Wenpin Tang\CU}
\maketitle

\begin{abstract}
Reinforcement Learning from human feedback (RLHF) has been shown a promising direction for aligning generative models with human intent and has also been explored in recent works \cite{DDPO, DPOK} for  alignment of diffusion generative models. In this work, we provide a rigorous treatment by formulating the task of fine-tuning diffusion models, with reward functions learned from human feedback, as an exploratory continuous-time stochastic control problem. 
Our key idea lies in treating the score-matching functions as controls/actions, and upon this, we develop a unified framework from a continuous-time perspective, to employ reinforcement learning (RL) algorithms in terms of improving the generation quality of diffusion models. We also develop the corresponding continuous-time RL theory for policy optimization and regularization under assumptions of stochastic different equations driven environment. 
Experiments on the text-to-image (T2I) generation will be reported in the accompanied paper \cite{ZC+}.
\bigskip

\end{abstract}

\section{Introduction}
Diffusion models \cite{sohl2015deep} can shape a simple noisy/non-informative initial distribution step-by-step into a complex target distribution by a learned denoising process \cite{Ho20DDPM,DDIM,Song20SGMbySDE} and have shown the remarkable capability to capture intricate, high-dimensional distributions make them the leading framework for high-quality and creative image generation, both unconditionally \cite{dhariwal2021diffusion} or conditionally given additional text prompts \cite{DALLE2, StableDiffusion, Imagen, DALLE3}; they are also rapidly finding use in other domains such as video synthesis \cite{ho2022imagen-video}, drug design \cite{xu2022geodiff} and continuous controls \cite{janner2022planning,wang2022diffusion}. However, existing models still have limited abilities for needs like multiple objective compositions \cite{feng2022training,gokhale2022benchmarking}, specific color and counts \cite{lee2023aligning}, and they may also suffer from sources of bias or fairness concern \cite{luccioni2023stable} and may fail to produce reliable visual text, even being distorted. As such, there is great interest in improving them further in terms of either generated distribution quality, or controllability. Specifically, due to the emergence of human-interactive platforms like ChatGPT \cite{ouyang2022training} and Stable Diffusion \cite{StableDiffusion}, there is increasing and substantial demand to let the generative models align with the user/human preference or feedback, which seems to be infeasible for the current training process only targeted at maximizing the likelihood.
\vskip 7 pt
Inspired by such needs, \cite{hao2022optimizing} proposed a natural way to fine tune or optimize diffusion models: Reinforcement Learning (RL, \cite{sutton2018reinforcement}). RL has already shown empirical success in enhancing Large Language Models' performance using human feedback \cite{bubeck2023sparks, christiano2017deep, ouyang2022training}; inspired by this, \cite{fan2023optimizing} is among the first to utilize RL-like methods to train diffusion
models for better image synthesis; \cite{DDPO, DPOK, lee2023aligning} further improve the text-to-image(T2I) diffusion model performance along by incorporating the reward model (like CLIP, BLIP, ImageReward which reflects human preference) under the lens of RL. The RL methods not only adapt to non-differentiable reward models, and are also shown in\cite{DDPO} to outperform benchmark supervised fine-tuning methods, like Reward Weighted Regression (RWR) \cite{lee2023aligning}, either with regularization or not \cite{DPOK}. Notably, all references above, which combine diffusion models with RL, are formulated as {\it discrete-time} sequential optimization problems, such as discrete-time Markov decision processes (MDPs, \cite{puterman2014markov}), and further solve the finite horizon MDPs by discrete-time RL algorithms like REINFORCE \cite{sutton1999policy} or PPO \cite{schulman2017proximal}. 
\vskip 7pt
However, the nature of the evolution of the diffusion models is essentially {\it continuous-time} \cite{Song20SGMbySDE}, as they are inspired by non-equilibrium thermodynamics \cite{sohl2015deep}. The continuous-time formulation of diffusion models provides a unified view on various existing discrete-time algorithms: to name a representative two, as shown by \cite{Song20SGMbySDE}, the discrete denoising steps inherent in DDPM \cite{Ho20DDPM} can be viewed as a discrete approximation to a stochastic differential equation (SDE, \cite{KS91}) and are implicitly {\it score-based} under a specific VP SDE (\cite{Song20SGMbySDE}); DDIM \cite{DDIM}, underpinning the success of Stable Diffusion \cite{StableDiffusion}, can also be seen as a numerical integrator of an Ordinary Differential Equation (ODE) sampler \cite{salimans2022progressive}. The awareness of the continuous-time nature, improves the design structure of current discrete-time SOTA large-scale text-to-image generative models e.g., in \cite{dhariwal2021diffusion,StableDiffusionv3,StableDiffusion}, and enables simple controllable generations by classifier guidance and solving inverse problems \cite{song2021solving,Song20SGMbySDE}. It also motivates more efficient diffusion models with continuous-time samplers, including ODE governed probability (normalizing) flows \cite{papamakarios2021normalizing,Song20SGMbySDE} and Rectified Flows \cite{RectifiedFlow,InstaFlow} underpinning Stable Diffusion v3 \cite{StableDiffusionv3} to achieve efficient generations. A discrete-time formulation of RL algorithms for tuning diffusion models are not directly applicable to continuous-time samplers; in addition, if directly applied to continuous-time diffusion models, it can {\it omit} the continuous nature and fail to capture or utilize the structural property: these call for a {\it continuous-time RL} approach. 
\vskip 7 pt
\noindent
This work aims to provide a framework for developing 
systematic approaches to improving RL algorithms efficiency for diffusion models alignment, and especially addressing the issue that current discrete-time frameworks are not applicable to deterministic samplers like DDIM or Rectified Flow, or any classical continuous-time SDE solvers. 
\begin{itemize}
\item
Our first contribution is a unified continuous-time framework for {\it diffusion models alignment}, by treating the score functions as actions. Thus the diffusion generative model of our interest is in the context of {\it score-based} models. Our framework thus allows continuous-time samplers from the nature of design and our formulations are also general enough to include either stochastic or deterministic samplers. 
\item
Our second contribution is the theory for policy optimization for continuous-time reinforcement learning, under the assumptions of underlying environments being an SDE (or ODEs when degenerate). Motivated by the downstream task of diffusion model alignment, our theory is centered around the case of maximizing a lump-sum reward with regularization, and results in novel continuous-time RL algorithms including continuous-time Policy Gradient, continuous-time version of PPO and GAE for the finite horizon. 
\end{itemize}
Empirical results and further development of the proposed approach will be reported in our forthcoming paper \cite{ZC+}.

\subsection{More Related works}
Our work is related to papers in several following domains.
\vskip 7pt
\textbf{Continuous-time reinforcement learning}. Recent years have witnessed a fast-growing body of research that has extended the frontiers of continuous-time RL in several important directions including, for instance, modeling the noise or randomness in the environment dynamics as following a SDE, and incorporating an entropy-based regularizer into the objective function \cite{wang2020reinforcement} to facilitate the exploration-exploitation tradeoff; designing model-free methods and algorithms under either finite horizon \cite{jia2022policy_evaluation, jia2022policy_gradient, jia2022q_learning} or infinite horizon \cite{zhao2024policy}, along with applications to portfolio optimization 
\cite{dai2023learning, huang2022achieving}.
\vskip 7 pt
\textbf{RL for fine-tuning T2I diffusion models}. DDPO \cite{DDPO} and DPOK \cite{DPOK} both discrete the time steps and finetune large pretrained text-to-image diffusion models through the reinforcement learning algorithms.
\vskip 7 pt
\textbf{Concurrent works}. Concurrent works of ours include \cite{uehara2024continuous-fine-tune}, which also formulated the diffusion models alignment as a continuous-time stochastic control problem with a different parameterization of the control; \cite{Tang24,uehara2024understanding} also provide a more rigorous review and discussion.
\cite{uehara2024feedback} investigated on how to learn the reward model in an online manner, based on the same stochastic control formulation.
\section{Formulation and Preliminaries}
\label{sc2}
In this section, we first review some necessary backgrounds of score-based diffusion models \cite{Song20SGMbySDE} (see \cite{SBDM_tutorial} for a tutorial and more comprehensive review) and continuous-time RL \cite{wang2020reinforcement}.
\subsection{Score Based Diffusion Models}

\textbf{Diffusion Models}. Consider perturbed data distributions evolve according to an SDE as the noise intensifies \cite{Song20SGMbySDE}, in which the forward SDE with state space $X_t\in \mathbb{R}^d$ is defined as:
\begin{equation}
\label{eq:SDE}
\mathrm{d}X_t = f(t, X_t) \mathrm{d}t + g(t) \mathrm{d}B_t, \quad \mbox{with } X_0 \sim p_{\scalebox{0.5}{data}}(\cdot),
\end{equation}
where $(B_t, \, t \ge 0)$ is $d$-dimensional Brownian motion, 
and $f: \mathbb{R}_+ \times \mathbb{R}^d \to \mathbb{R}^d$ 
and $g: \mathbb{R}_+ \to \mathbb{R}_+$
are designed model parameters. To avoid technical difficulties, we assume that the stochastic processes \eqref{eq:SDE} are well-defined, see e.g., \cite[Section 5.3]{KS91} or \cite[Chapter 6]{SV79} for background. Denote $p(t, \cdot)$ as the probability density of $X_t$.
\vskip 7pt
Set time horizon $T > 0$ to be fixed, and run the SDE \eqref{eq:SDE} until time $T$ to get $X_T \sim p(T, \cdot)$. The time reversal $X^{\text{rev}}_t: = X_{T-t}$ for $0 \le t \le T$, such that $\operatorname{law}(X^{\text{rev}}_t)=\operatorname{law}(X_{T-t})$ for every $t\in[0,T]$, also satisfies an SDE, under some mild conditions on $f$ and $g$:
\begin{equation}
\label{eq:timerev_SDE}
\mathrm{d} X^{\text{rev}}_t = \left(-f(T-t,X^{\text{rev}}) + \frac{1+\eta^2}{2}g^2(T-t) \nabla_x \log p(T-t, X^{\text{rev}})  \right) \mathrm{d}t + \eta g(T-t) \mathrm{d}B_t,
\end{equation}
in which $\nabla_x \log p(t, x)$ is known as the stein score function \cite{Song20SGMbySDE}) and $\eta\in[0,1]$ is a constant. If starting from the distribution $p(T,\cdot)=\operatorname{law}(X_T)$, the SDE \eqref{eq:timerev_SDE} runs until time $T$ and will generate (or recover) the desired $X^{\text{rev}}_T  \sim p_{\scalebox{0.5}{data}}(\cdot)$. However, since distribution $p(T, \cdot)$ is unknown and hard to sample from, diffusion models typically replace it with a prior distribution $p_\infty(\cdot)$ such that $p(T, \cdot)\approx p_\infty(\cdot)$ when $T$ is sufficiently large (which can motivate a smart choice of diffusion dynamics, see e.g. \cite{CDPM} for some discussions.). The prior distribution is thus independent of $p(0, \cdot) = p_{\scalebox{0.5}{data}}(\cdot)$, and this explains why DPMs generate distributions from ``noise'' $p_\infty(\cdot)$. In addition, a special but important case by taking $\eta = 0$ in the \eqref{eq:timerev_SDE}, this results to an ODE \cite{Song20SGMbySDE}:
\begin{equation}
\label{eq:timerev_ODE}
\mathrm{d} X^{\text{ode}}_t = \left(-f(T-t,X^{\text{ode}}) + \frac{1}{2}g^2(T-t) \nabla_x \log p(T-t, X^{\text{ode}})  \right) \mathrm{d}t,
\end{equation}
which can enable faster sampling and likelihood computation thanks to the deterministic generation.
\vskip 7pt
\textbf{Score Matching}. Since the score function $\nabla_x \log p(t, x)$ is also unknown, diffusion models learn a function approximation $s_{\theta}(t,x)$, parameterized by $\theta$ (usually a neural network), to the true score by minimizing the MSE or the Fisher divergence (between the learned distribution and true distribution), evaluated by samples generated through the forward process \eqref{eq:SDE}:
\begin{equation}
\label{eq:SM objective}
\mathcal{J}(\theta) = \mathbb{E}_{t\sim\operatorname{Uni}(0,T)}\mathbb{E}_{x\sim p(t,\cdot)}\left\{\lambda(t) \left[\left\|s_{\theta}(t,x)-\nabla_{x} \log p(t,x)\right\|_2^2\right]\right\},
\end{equation}
in which $\lambda:[0, T] \rightarrow \mathbb{R}_{>0}$ is a chosen positive weighting function. When choosing the weighting functions as $\lambda(t) = g^2(t)$, this score matching objective is equivalent to maximizing an evidence lower bound (ELBO, \cite{song2021maximum,huang2021variational}) of the log-likelihood. The explicit score matching (ESM) objective \eqref{eq:SM objective} is equivalent to the following denoising score matching (DSM) \cite{vincent2011connection} objective:
\begin{equation}
\label{eq:DSM objective}
\theta^*=\arg\max_{\theta}\mathbb{E}_{t\sim\operatorname{Uni}(0,T)}\left\{\lambda(t) \mathbb{E}_{x_0\sim p_{data}} \mathbb{E}_{x_t \mid x_0}\left[\left\|s_{\theta}(t,x_t)-\nabla_{x_t} \log p(t,x_t \mid x_0)\right\|_2^2\right]\right\}
\end{equation}
in which $x_t \sim p(t,\cdot|x_0)$. DSM objective is tractable and also easier to be optimized in the sense that $p(t,\cdot|x_0)$ can be written in closed-form given a smart choice of diffusion dynamics, for example, it is conditionally Gaussian if \eqref{eq:SDE} is a linear SDE as in \cite{Song20SGMbySDE}.
\vskip 7pt
After learning the optimal approximation $s_{\theta^*}$ which minimizes the objective, diffusion models replace the true score $\nabla \log p(t, x)$ by the learned score $s_{\theta^*}(t,x)$ in \eqref{eq:timerev_SDE} as an approximation to $X^{\text{rev}}$, which we denotes as $X^{\sla}_t$, such that:
\begin{equation}
\label{eq:timerevapprox}
\mathrm{d} X^{\sla}_t = \left(-f(T-t,\overline{X}_t) + \frac{1+\eta^2}{2}g^2(T-t) s_{\theta^{*}}(T-t, X^{\sla}_t)  \right) \mathrm{d}t + \eta g(T-t) \mathrm{d}B_t,\ X^{\sla}_0\sim p_\infty(\cdot),
\end{equation}
at time $T$ to achieve generation, or the corresponding ODE sampler ($\eta = 0$) as:
\begin{equation}
\label{eq:timerevapprox_ode}
\mathrm{d} X^{\sla}_t = \left(-f(T-t,X^{\sla}_t) + \frac{1}{2}g^2(T-t) s_{\theta^{*}}(T-t, X^{\sla}_t)  \right) \mathrm{d}t,\ X^{\sla}_0\sim p_\infty(\cdot).
\end{equation}
The well-known DDPM \cite{Ho20DDPM} sampler and DDIM \cite{DDIM} sampler can both be seen as a certain discretization/integration role of the SDE in \eqref{eq:timerevapprox} or the ODE in \eqref{eq:timerevapprox_ode} as proved by \cite{salimans2022progressive, Song20SGMbySDE, zhang2022gddim, zhang2009exponential}, we include the more detailed discussions in Appendix \ref{app:discrete and continuous sampler connection} for completeness. We will follow the continuous-time prospective throughout this paper. 
\vskip 7pt
For the backward process \eqref{eq:timerevapprox} and \eqref{eq:timerevapprox_ode}, the score function approximations $s_{\theta^{*}}(T-t, X^{\sla}_t)$ determine the $\operatorname{law}(X^{\sla}_t)$s throughout the backward process with $t\in(0,T]$, especially $\operatorname{law}(X^{\sla}_T)$ which we may actually be interested in. This is reminiscent of the actions or controls taken by agents in the RL/control theory and our formulation are thus motivated to treat this score function/approximation as a {\it control} over the backward process; for task of fine-tuning, $s_{\theta^{*}}$ can be seen as a pretrained control/policy we can both have access and we shall refer to.

\subsection{Continuous-time RL}
We start with a general formulation of the continuous-time RL under a finite horizon as a stochastic control problem, based on the same modeling framework as in \cite{jia2022policy_evaluation, wang2020reinforcement}.
\vskip 7pt
\textbf{Diffusion Process.} Assume that the state space is $\mathbb{R}^d$, and denote by $\mathcal{A}$ the action space. 
Let $\pi(\cdot \mid t,x) \in \mathcal{P}(\mathcal{A})$ be a {(state) feedback} policy given the time $t\in [0,T]$ and state $x \in \mathbb{R}^d$. 
A continuous RL problem is formulated by a distributional (or relaxed) control approach \cite{yong1999stochastic}, 
which is motivated by the trial and error process in RL. 
The state dynamics $(X^a_t, \, t \ge 0)$ is governed by the 
It\^o process:
\begin{equation}
\label{SDE_dynamic}
\mathrm{d} X_t^a=b\left(t, X_t^a, a_t\right) \mathrm{d} t+\sigma\left(t\right) \mathrm{d} B_t,\quad X^a_0\sim \rho\in\mathcal{P}(\mathbb{R}^d),
\end{equation}
where $(B_t, \, t \ge 0)$ is the $m$-dimensional Brownian motion,
$b:[0,T]\times\mathbb{R}^d \times \mathcal{A} \mapsto \mathbb{R}^d$,
$\sigma:[0,T]\mapsto \mathbb{R}_+$,
and the action $a_s$ is generated from the distribution $\pi\left(\cdot \mid t, X^a_t\right)$ by {\em external randomization}.
\vskip 7pt
From now on, write $(X^{\pi}_t, a^{\pi}_t)$ for the state and action at time $t$ given by the process \eqref{SDE_dynamic} under the policy $\pi = \{\pi(\cdot \mid t, x) \in \mathcal{P}(\mathcal{A}): t \in [0,T], x \in \mathbb{R}^d\}$. 
The goal here is to find the optimal feedback policy {$\pi^*$ that maximizes the expected reward over a finite time horizon:}
\begin{equation}
\label{Discounted Objective 2}
{V^*: =} \max_{\pi} \mathbb{E}\left[\int_0^{T}\left[r\left(t, X_t^\pi, a_t^\pi\right)+\gamma R\left(t, X_t^\pi, a_s^\pi, \pi\left(\cdot \mid X_t^\pi\right)\right)\right] \mathrm{d} t +h(X^{\pi}_T)\mid X_0^\pi\sim \mu\right],
\end{equation}
where $r:[0,T]\times\mathbb{R}^d \times \mathcal{A} \mapsto \mathbb{R}^{+}$ is the running reward of {the current state and action $(X^\pi_s, a^\pi_s)$}; $h:\mathbb{R}^d \mapsto \mathbb{R}^{+}$ being the lump-sum reward function applied at the end of the period $T$; $R: [0,T]\times\mathbb{R}^n \times \mathcal{A} \times \mathcal{P}(\mathcal{A}) \mapsto \mathbb{R}$ is a regularizer {which facilitates exploration} (e.g., in \cite{wang2020reinforcement}, $R$ is taken as the differential entropy defined by $R(t, x, a, \pi(\cdot))=-\log \pi(a)$);
$\gamma \geq 0$ is a weight parameter on exploration (also known as the ``temperature'' parameter). 
\vskip 7pt
\textbf{Performance metric}. Given a policy $\pi(\cdot)$, let $\tilde{b}(t,x, \pi(\cdot)):=\int_{\mathcal{A}} b(t,x, a) \pi(a) \mathrm{d} a$, and consider the following equivalent SDE representation of \eqref{SDE_dynamic}:

\begin{equation}
\label{SDE_Dynamics_exp}
\mathrm{d} \tilde{X}_t = \tilde{b}\left(t,\tilde{X}_t, \pi(\cdot \mid  \tilde{X}_t)\right) \mathrm{d} t+\sigma\left(t\right) \mathrm{d} \tilde{B}_t, \quad \tilde{X}_0\sim \rho,
\end{equation}
in the sense that
there exists a probability measure $\tilde{\mathbb{P}}$ 
which supports the $d$-dimensional Brownian motion $(\tilde{B}_s, \, s \ge 0)$, 
and for each $s \geq 0$, the distribution of $\tilde{X}_s$ under $\tilde{\mathbb{P}}$ agrees with that of $X_s$ under $\mathbb{P}$ defined by \eqref{SDE_dynamic}. Note that the dynamics in \eqref{SDE_Dynamics_exp} does not require external randomization. We also set
$\tilde{r}(t,x,\pi):=\int_{\mathcal{A}} r(t,x, a) \pi(a) \mathrm{d} a$
and $\tilde{R}(t,x,\pi):=\int_{\mathcal{A}} R(t,x, a,\pi) \pi(a) \mathrm{d} a$.
\vskip 7pt
We formally define the (state) value function {given the feedback policy $\{\pi(\cdot \mid x): x \in \mathbb{R}^d$\}} by $V( t, x ; \pi):=$
\begin{equation}
\label{Value function Definition}
\begin{aligned}
&\mathbb{E} \left[\int_t^{T} \left[r\left(s, X_s^\pi, a_s^\pi\right)+\gamma R\left(s, X_s^\pi, a_s^\pi, \pi\left(\cdot \mid X_s^\pi\right)\right)\right] \mathrm{d} s +h\left(X_T^{\pi}\right)\mid X_0^\pi = x \right] \\
=\,& \mathbb{E} \left[\int_t^{T}  \left[\tilde{r}\left(s,\tilde{X}_s^\pi, \pi(\cdot \mid \tilde{X}_s^\pi)\right)+\gamma \tilde{R}\left(s,\tilde{X}_s^\pi ,\pi(\cdot \mid \tilde{X}_s^\pi)\right)\right] \mathrm{d} s+ h\left(\tilde{X}_T^{\pi}\right)\mid \tilde{X}_0^\pi=x\right],
\end{aligned}
\end{equation}
and the performance metric as: $\eta(\pi):= \int_{\mathbb{R}^n} V(0, x; \pi) \mu(dx)$,
so $V^* = \max_\pi \eta(\pi)$. 
The main task of the continuous RL is to approximate
$\max_\pi \eta(\pi)$ by constructing a sequence of policies $\pi_k$, $k = 1,2,\ldots$ recursively such that $\eta(\pi_k)$ is non-decreasing.
\vskip 7pt
It will also be useful to define $q$-value function \cite{jia2022q_learning} as: define the $q$-value as 
\begin{equation}
\label{defqvalue}
q(t, x, a ; \pi):=\frac{\partial V}{\partial t}\left(t, x ; \pi\right)+\mathcal{H}\left(t, x, a, \frac{\partial V}{\partial x}\left(t,x ; \pi\right), \frac{\partial^2 V}{\partial x^2}\left(t,x ; \pi\right)\right),
\end{equation}
for a given policy $\pi \in \Pi$ and $(t, x, a) \in[0,T]\times\mathbb{R}^d \times \mathcal{A}$, in which $\mathcal{H}(t, x, a, y, A):=b(t, x, a) \cdot y+\frac{1}{2} \sigma^2(t)\circ A+r(t, x, a)$ is the (generalized) Hamilton function in stochastic control theory \cite{yong1999stochastic}. 

\section{Main Results}
\label{sc3}
\subsection{Fine-tuning Diffusion Models as Stochastic Control}
\label{sec: Aligning Stochastic Sampler}
In this section, we formally formulate the task of fine-tuning diffusion models as a continuous-time exploratory stochastic control problem under the finite horizon. The key component under our formulation is to regard the backward process as a stochastic control process and treat the score function approximation as action/control process.
\vskip 7 pt
\textbf{Score function as action}. The backward procedure of diffusion models as defined in \eqref{eq:timerevapprox}, when we generate conditionally on an additional input/context which we denote as $c$. $c$ can be a label for generate imaging of a certain class or an input text prompt for text-to-image generation, and the reverse process in the (latent) diffusion models becomes
\begin{equation}
\mathrm{d} X^{\sla}_t = \left(-f(T-t, X^{\sla}_t) + \frac{1+\eta^2}{2}g^2(T-t) s_{\theta_{\text{pre}}}(T-t, X^{\sla}_t,c)\right) \mathrm{d}t + \eta g(T-t) \mathrm{d}B_t,
\end{equation}
in which we abuse the notation of $s_{\theta_{\text{pre}}}(T-t,  X^{\sla}_t,c)$ to represent the pretrained score function conditioning on $c$. Note that if we choose the $b$ and $\sigma$ in the continuous RL dynamics assumption in \eqref{SDE_Dynamics_exp} as the diffusion term being $\sigma(t)=g(T-t)$, and the drift term being:
\begin{equation}
\label{eqn:drift as action}
b_{\eta}\left(t, x, a\right) := -f(T-t,x) + \frac{1+\eta^2}{2} g^2(T-t) a,
\end{equation}
then if we define a specific (deterministic) feedback control process as $a^{\text{pre}}_t= s_{\theta_{\text{pre}}}(T-t,  X^{\sla}_t,c)$, the backward procedure can be rewritten as:
\begin{equation}
\label{eq:score as action}
\mathrm{d} X^{\sla}_t = b_{\eta}\left(t, X^{\sla}_t, a_t\right)\mathrm{d}t + \eta \sigma(t) \mathrm{d}B_t,
\end{equation}
thus the action has a clear meaning as the place of score function (conditional on the input context $c$). Thus we treat the current score approximation as a pretrained feedback control/action process, which we want to further optimize on for diffusion alignment.
\vskip 5 pt
\textbf{Exploratory SDEs}. Lies central in RL is the exploration. To encourage exploration, at time $t$, we can adopt a Gaussian exploration control/policy 
$$
a^{\theta}_t \sim \pi^{\theta}(\cdot\mid t,X_t,c) = N(\mu^{\theta}(t,X_t,c),\Sigma_t\cdot I),
$$
in which the mean $\mu^{\theta}(t,X_t,c)$ is approximated by some function approximation parameterized by $\theta$ and the variance $\Sigma_t$ is a chosen fixed exploration level for each $t$. We denote the corresponding process $X^{\pi^{\theta}}_t$ as $X^{\theta}_t$ for brevity. Under this Gaussian parameterization, similar to discussion of \eqref{SDE_Dynamics_exp}, it suffices to consider the following SDE (it actually holds for general distribution beyond Gaussian with mean $\mu^{\theta}(t,X_t,c)$) governed by a corresponding deterministic policy only dependent on the mean $\mu^{\theta}(t,X_t,c)$:
\begin{equation}
\mathrm{d} X_t^{\theta}=\left[-f(T-t,X_t^{\theta}) + \frac{1+\eta^2}{2}g^2(T-t) \mu^{\theta}(t,X_t^{\theta},c)\right]\mathrm{d}t+\eta g(T-t)\mathrm{d}t,\quad X^{\theta}_0\sim\rho,
\end{equation}
and we denote the probability density of $X_t^{\theta}$ as $p_{\theta}(t,\cdot,c)$. 
\vskip 5 pt 
We emphasize that using the (conditional) score matching parameterization as the same as the policy parameterization $\mu^{\theta^*}(t,x,c) = s^{\theta^*}(T-t,x,c)$ recovers the backward procedure \eqref{eq:timerevapprox} in the score-based diffusion models, that's why we refer our formulation enables `score function as action'. This perspective is also reminiscent to classifier guidance, as classifier guidance for diffusion models leverage the property of the conditional score as:
$$
\nabla \log p(t,x\mid c) = \nabla \log p(t,c\mid x)+\nabla \log p(t,x),
$$
in order to generate conditionally on an extra label $c$ in our contents. $\nabla \log p(t,x)$ can be seen as an old unconditional control before fine-tuned, and $\nabla \log p(t,x\mid y)$ can be seen as a new control for better conditional generation, and $ \nabla p(T,c\mid x)$ acts as a reward to guide the shift/difference from the old control to the new one. In more general cases when reward model is complicated, it's hard to design structures like the case of classifier guidance, for which $\nabla \log p(t,y\mid x)$ appears as a natural candidate for conditional likelihood maximization $ p(T,y\mid x)$, that's why we use the RL approach.

\vskip 7 pt
\textbf{Regularization as Rewards}. We assume that we are given a reward model (RM), such that it can output the reward $\text{RM}(x,c)$ given a generation $x\in\mathbb{R}^d$, which represents the human preference or any target we want to maximize of the current generation $X_T$ given input $c$. For example, if the downstream task is text-to-image generation, we have $\text{RM}(x,c)$ represents how the generated image $X_T$ align with input prompt $c$. In addition, we also consider adding regularization: in our cases, there should be two sources of regularization: (i) regularization to encourage exploration as in our continuous RL formulation, like entropy regularization in \cite{wang2020reinforcement} or \cite{jia2022policy_evaluation}; (ii) regularization to prevent the model from overfitting to the reward or catastrophic forgetting, and thus failing to utilize the capabilities of the pre-trained diffusion models parameterized by $\theta_{pre}$. Notice that here since we already used Gaussian Exploration with fixed variance to encourage exploration, we will not use additional regularization of source (i); for source (ii), in the same essence of previous work of tuning diffusion models by discrete-time RL \cite{DPOK, ouyang2022training}, we still target at bounding the KL divergence of the final generation, i.e., our final optimization objective yields:
\begin{equation}
\label{objective with regularization}
\mathbb{E}\left[\text{RM}(c,X^{\theta}_T)-\beta \operatorname{KL}(p_{\theta}(T,\cdot,c)\|p_{\theta_{pre}}(T,\cdot,c))\right],
\end{equation}
in which $\beta$ is a penalty constant, often needed to tune separately. However, instead of adding KL divergence directly to the objective like DPOK \cite{DPOK}, we could `smartly' transform this term into an integration of expected $L^2$ penalty term between the mean of the current Gaussian policy and the pretrained/reference score along the path, thanks to our continuous-time formulation and the following theorem:
\begin{thm}
\label{thm:Regularization as KL bound}
We have that, for any $c$, the discrepancy between the $p_{\theta}$ and $p_{\theta_{pre}}$ satisfies:
\begin{equation}
\operatorname{KL}(p_{\theta}(T,\cdot,c)\|p_{\theta_{pre}}(T,\cdot,c))= \frac{1}{2}\left(\frac{1+\eta^2}{2\eta}\right)^2\int_{0}^{T}\mathbb{E}_{p_{\theta}(t,\cdot\mid c)} g^2(T-t)\|\mu^{\theta}(t,X_t^{\theta},c)-\mu^{\theta_{pre}}(t,X_t^{\theta},c)\|^2\mathrm{d}t.
\end{equation}
\end{thm}
The proof is enclosed in Appendix \ref{Proof of Regularization as KL bound}. As a remark, it is important to let the expectation under $p^{\theta}$ in order for later online update. In addition, notice that
$$
\mathbb{E}_{p_{\theta}(t,\cdot)}\|a_t^{\theta}-\mu^{\theta_{pre}}(t,X_t,c)\|^2=\mathbb{E}_{p_{\theta}(t,\cdot\mid c)} \|\mu^{\theta}(t,X_t^{\theta},c)-\mu^{\theta_{pre}}(t,X_t^{\theta},c)\|^2+ \Sigma_t,
$$
we have that optimizing \eqref{objective with regularization} is equivalent to optimizing the following objective:
\begin{equation}
\begin{aligned}
\label{continuous-time objective with regularization}
\eta^{\theta}&=\mathbb{E}\int_0^{T}\underbrace{-\frac{\beta}{2}\left(\frac{1+\eta^2}{2\eta}\right)^2 g^2(T-t)\|a_t^{\theta}-\mu^{\theta_{pre}}(t,X^\theta_t,c)\|^2}_{r_{\eta}(t,X^\theta_t,a_t^{\theta})} \mathrm{d} t+\text{RM}(c,X^{\theta}_T)
\end{aligned}
\end{equation}
inspired by the previous Theorem \ref{thm:Regularization as KL bound}, and This aligns with our continuous-time RL formulation reviewed in \eqref{Discounted Objective 2}, and we further investigate the methodologies for policy optimization under this formulation.

 

We first present the policy gradient formula for our finite horizon problem \eqref{Discounted Objective 2}, which simplifies the formula in \cite{jia2022policy_gradient} by utilizing the discussion in \cite{jia2022q_learning,zhao2024policy}. 
\begin{thm}
\label{thm:PG formula}
We have that the policy gradient of an admissable policy $\pi^{\theta}$ parameterized by $\theta$ is:
\begin{equation}
\begin{aligned}
\nabla_{\theta}\eta^{\theta}= & \mathbb{E}^{\mathbb{P}}\left[\int _ { 0 } ^ { T }\nabla_{\theta} \log \pi^ { \theta} ( a _ { t } ^ {\theta} | t , X _ { t } ^ {\theta} ) q(t, X_t^{\theta}, a_t^{\theta} ; \pi^\theta)\mathrm{d} t\right],
\end{aligned}
\end{equation}
\end{thm}
The proof is enclosed in Appendix \ref{Proof of PG formula}. Further notice that the only terms in the $q$-value functions that are connected to the action $a$ are:
$$
r_{\eta}(t,x,a)+b(t,x,a)\cdot \frac{\partial V}{\partial x}\left(t,x ; \pi\right)=r_{\eta}(t,x,a)+(-f(T-t,x) + \frac{1+\eta^2}{2}g^2(T-t) a)\cdot \frac{\partial V}{\partial x}\left(t,x ; \pi\right),
$$
in which $r_{\eta}(t,x,a)$ can be observed and $b(t,x,a)$ can be computed given $x,a$. When the reward function is differentiable with respect to $x$, then $\frac{\partial V}{\partial x}\left(t,x ; \pi\right)$ can be evaluated by backward prorogation directly. Thus we can leverage this structural property and yield the CPG formula: 
\begin{cor}
\label{thm:PG formula simplified}
We have that the policy gradient of an admissable policy $\pi^{\theta}$ parameterized by $\theta$ is:
\begin{equation}
\label{SDE continuous-time PG}
\begin{aligned}
\nabla_{\theta}\eta^{\theta}= & \mathbb{E}^{\mathbb{P}}\left[\int _ { 0 } ^ { T }\nabla_{\theta} \log \pi^ { \theta} ( a _ { t } ^ { \theta } | t , X _ { t }^ {\theta} ) \left(
r_{\eta}(t,X _ { t } ^ {\theta},a _ { t } ^ {\theta})+\frac{1+\eta^2}{2}g^2(T-t) a _ { t } ^ { \theta} \cdot \frac{\partial V}{\partial x}\left(t,X _ { t } ^ {\theta} ; \pi\right)\right)\mathrm{d} t\right],
\end{aligned}
\end{equation}
\end{cor}
This observation allows us only to estimate the value function, which we call as the continuous-time case of Generalized Advantage Estimation (GAE) \cite{GAE}.

\subsection{Directly aligning deterministic sampler}
\label{sec: Aligning Deterministic Sampler}
We can also directly fine-tune an deterministic sampler instead of using the related SDEs. For example, Stable Diffusion v1 adopts DDIM, which can be seen as an integration role of ODE; Stable Diffusion v3 \cite{StableDiffusionv3} is built upon Rectified Flow \cite{RectifiedFlow,InstaFlow}. 
In this work, we specifically consider the case of fine-tuning an ODE-based model parameterized (by $\phi$) as:
\begin{equation}
\mathrm{d} X_t^{\phi}=(-f(T-t,X_t^{\phi}) + \frac{1}{2}g^2(T-t) \underbrace{\mu^{\phi}(t,X_t^{\phi},c)}_{a^{\phi}_t})\mathrm{d}t,\quad X^{\phi}_0\sim\rho.
\end{equation}
For brevity of analysis, we assume that the pre-trained model parameterized by $\phi_{pre}$ yields perfect score matching, i.e. $s^{\theta_{pre}}(t,x,c)=\nabla_x p(t,x,c)$ for any $t,x,c$; Moreover we consider $\rho=\mathcal{N}(0,I)$, which indeed holds for forward process being VP-SDE. We consider the exploratory ODE by taking $a^{\phi}_t\sim \mathcal{N}(\mu^{\phi}(t,X_t^{\phi},c),\sigma_t^2)$, similar to our exploratory SDE cases. To optimize the objective in \eqref{objective with regularization} in which $X_t$ now follows exploratory ODE (instead of SDE), we use the idea of MM algorithm, and then derive the policy gradient of a lower bound of the regularized objective:
\vskip 5 pt

\begin{thm} We have for any context $c$, there exists a constant $\tilde{C}$ such that:
\begin{equation}
\operatorname{KL}(p_{\phi^*}(\cdot,c)\|p_{\phi_{pre}}(\cdot,c))\leq \tilde{C} \int_{0}^{T}g^2(T-t)\|\mu_{\phi^*}(t,\cdot,c)-\mu_{\phi_{pre}}(t,\cdot,c)\|^2_{H^1}\mathrm{d}t
\end{equation}
\end{thm}
More details of the theorem will be developed in the accompanied paper \cite{ZC+}.
Notice that, we need this separate argument since directly letting $\eta\rightarrow 0$ provides a vacuous bound in Theorem \ref{thm:Regularization as KL bound}. This theorem thus motivates our following objective as:
\begin{equation}
\begin{aligned}
\label{continuous-time objective with regularization ODE}
\eta^{\phi}&=\mathbb{E}\int_0^{T}\underbrace{-\beta \tilde{C} g^2(T-t)\|a_t^{\phi}-\mu^{\phi_{pre}}(t,X^\phi_t,c)\|^2}_{r(t,X^\phi_t,a_t^{\phi})} \mathrm{d} t+r(c,X^{\phi}_T)
\end{aligned}
\end{equation}
for which we reduce the $H^1$ norm to $L^2$ norm for computation efficiency, and it is thus the same as in SDE case despite that the expectation is now under the exploratory ODE. For practical concern, since $\tilde{C}$ is unknown, $\beta \tilde{C}$ can be tuned together to obtain final good results.
\vskip 5 pt
\textbf{Continuous-time}. We can also derive the continuous-time policy gradient without prior time discretization as (which corresponds to Corollary \ref{thm:PG formula simplified} with $\eta=0$, though with different reward function definition):
\begin{cor}
\label{thm:PG formula ODE}
We have that the policy gradient of an admissable policy $\pi^{\phi}$ parameterized by $\phi$ is:
\begin{equation}
\begin{aligned}
\nabla_{\phi}\eta(\pi^{\phi})= & \mathbb{E}^{\mathbb{P}}\left[\int _ { 0 } ^ { T }\nabla_{\phi} \log \pi^ { \phi} ( a _ { t } ^ { \pi ^ {\phi} } | t , X _ { t } ^ {\phi } ) \left(r(t,X _ { t } ^ {\phi },a _ { t } ^ {\phi}) + b_{0}(t,X _ { t } ^ {\phi},a _ { t } ^ { \phi})\cdot \frac{\partial V}{\partial x}\left(t,x ; \pi\right)\right)\mathrm{d} t\right].
\end{aligned}
\end{equation}
\end{cor}

\section{Conclusions}
In this work, we proposed a continuous-time reinforcement learning framework for fine-tuning continuous-time diffusion models, guided by a pre-trained reward function (from human feedback). The work lays out the program; further development and experiments will be reported in forthcoming paper \cite{ZC+}.

\newpage

\bibliographystyle{myplainnat}
\bibliography{RL_diffusion}{}

\newpage

\begin{appendices}
\section{Connection between discrete-time and continuous-time sampler}
In this section, we summarize the discussion of popular samplers like DDPM, DDIM, stochastic DDIM and their continuous-time limits.
\label{app:discrete and continuous sampler connection}
\subsection{DDPM sampler is the discretization of VP-SDE}
\label{app:ddpm}
We review the forward and backward process in DDPM, and its connection to the variance preserving (VP) SDE following the discussion in \cite{Song20SGMbySDE,SBDM_tutorial}. DDPM considers a sequence of positive noise scales $0<\beta_1, \beta_2, \cdots, \beta_N<1$. For each training data point $x_0 \sim p_{\text {data }}(x)$, a discrete Markov chain $\left\{x_0, x_1, \cdots, x_N\right\}$ is constructed such that:
\begin{equation}
\label{DDPM forward}
x_i=\sqrt{1-\beta_i} x_{i-1}+\sqrt{\beta_i} z_{i-1}, \quad i=1, \cdots, N,
\end{equation}
where $z_{i-1} \sim \mathcal{N}(0, I)$, thus $p\left(x_i \mid x_{i-1}\right)=\mathcal{N}\left(x_i ; \sqrt{1-\beta_i} x_{i-1}, \beta_i I\right)$. We can further think of $x_i$ as the $i^{\text{th}}$ point of a uniform discretization of time interval $[0,T]$ with discretization stepsize $\Delta t=\frac{T}{N}$, i.e. $x_{i \Delta t}=x_i$; and also $z_{i \Delta t}=z_i$. To obtain the limit of the Markov chain when $N \rightarrow \infty$, we define a function $\beta:[0,T]\rightarrow \mathbb{R}^+$ assuming that the limit exists: $\beta(t)= \lim_{\Delta t\rightarrow 0}\beta_i /\Delta t$ with $i=t/\Delta t$. Then when $\Delta t$ is small, we get:
$$x_{t+\Delta t}\approx\sqrt{1-\beta(t) \Delta t} x_t+\sqrt{\beta(t) \Delta t} z_t \approx x_t-\frac{1}{2} \beta(t) x_t \Delta t+\sqrt{\beta(t) \Delta t} z_t.$$
Further taking the limit $\Delta t \rightarrow 0$, this leads to:
$$
d X_t=-\frac{1}{2} \beta(t) X_t d t+\sqrt{\beta(t)} d B_t, \quad 0 \leq t \leq T,
$$
and we have:
$$
f(t, x)=-\frac{1}{2} \beta(t) x, g(t)=\sqrt{\beta(t)}.
$$
Through reparameterization, we have $p_{\bar{\alpha}_i}\left(x_i \mid x_0\right)=\mathcal{N}\left(x_i ; \sqrt{\bar{\alpha}_i} x_0,\left(1-\bar{\alpha}_i\right) I\right)$, where $\bar{\alpha}_i:=\prod_{j=1}^i\left(1-\beta_j\right)$. For the backward process, a variational Markov chain in the reverse direction is parameterized with $p_{\theta}\left(x_{i-1} \mid x_i\right)=\mathcal{N}\left(x_{i-1} ; \frac{1}{\sqrt{1-\beta_i}}\left(x_i+\beta_i s_{\theta}\left(i,x_i\right)\right), \beta_i I\right)$, and trained with a re-weighted variant of the evidence lower bound (ELBO):
$$
\theta^*=\underset{\theta}{\arg \min } \sum_{i=1}^N\left(1-\bar{\alpha}_i\right) \mathbb{E}_{p_{\text {data }}(x)} \mathbb{E}_{p_{\bar{\alpha}_i}(\tilde{x} \mid x)}\left[\left\|s_{\theta}(i,\tilde{x})-\nabla_{\tilde{x}} \log p_{\bar{\alpha}_i}(\tilde{x} \mid x)\right\|_2^2\right] .
$$
After getting the optimal model $s_{\theta^*}(i,x)$, samples can be generated by starting from $x_N \sim \mathcal{N}(0, I)$ and following the estimated reverse Markov chain as:
\begin{equation}
\label{DDPM Backward Process}
x_{i-1}=\frac{1}{\sqrt{1-\beta_i}}\left(x_i+\beta_i s_{\theta^*}\left(i,x_i\right)\right)+\sqrt{\beta_i} z_i, \quad i=N, N-1, \cdots, 1 .
\end{equation}
Similar discussion as for the forward process, the equation \eqref{DDPM Backward Process} can further be rewritten as:
\begin{equation}
\begin{aligned}
x_{(i-1)\Delta t} &\approx \frac{1}{\sqrt{1-\beta_{i\Delta t}\Delta t}}\left(x_{i\Delta t}+\beta (i\Delta t)\Delta t \cdot s_{\theta^*} \left(i\Delta t,x_{i\Delta t}\right)\right)+\sqrt{\beta_i} z_i,\\
&\approx (1+\frac{1}{2}\beta_{i\Delta t}\Delta t)\left(x_{i\Delta t}+\beta (i\Delta t)\Delta t \cdot s_{\theta^*} \left(i\Delta t,x_{i\Delta t}\right)\right)+\sqrt{\beta_i} z_i,\\
&\approx (1+\frac{1}{2}\beta_{i\Delta t}\Delta t)x_{i\Delta t}+\beta (i\Delta t)\Delta t\cdot s_{\theta^*} \left(i\Delta t,x_{i\Delta t}\right)+\sqrt{\beta_i} z_i,
\end{aligned}
\end{equation}
when $\beta_{i\Delta t}$ is small. This is indeed the time discretization of the backward SDE:
\begin{equation}
\begin{aligned}
\mathrm{d} X^{\sla}_t &= (\frac{1}{2} \beta(T-t) X^{\sla}_t + \beta(T-t) s_{\theta^{*}}(T-t, X^{\sla}_t))\mathrm{d}t + \sqrt{\beta(t)} \mathrm{d}B_t,\\
&= \left(-f(T-t,X^{\sla}_t) + g^2(T-t) s_{\theta^{*}}(T-t, X^{\sla}_t)  \right) \mathrm{d}t + g(T-t) \mathrm{d}B_t.
\end{aligned}
\end{equation}

\subsection{DDIM sampler is the discretization of ODE}
\label{app:ddim}
We review the backward process in DDIM, and its connection to the probability flow ODE following the discussion in \cite{Song20SGMbySDE,kingma2021variationalDM,salimans2022progressive,zhang2022fast}.
\vskip 5 pt
(i) DDIM update rule: 
The concrete updated rule in DDIM paper (same as in the implementation) adopted the following rule (with $\sigma_t=0$ in Equation (12) of \cite{DDIM}):
\begin{equation}
\label{eq: DDIM update rule}
\boldsymbol{x}_{t-1}=\sqrt{\bar{\alpha}_{t-1}} \underbrace{\left(\frac{\boldsymbol{x}_t-\sqrt{1-\bar{\alpha}_t} \epsilon_\theta^{(t)}\left(\boldsymbol{x}_t\right)}{\sqrt{\bar{\alpha}_t}}\right)}_{\text {``predicted } \boldsymbol{x}_0 "}+\underbrace{\sqrt{1-\bar{\alpha}_{t-1}} \cdot \epsilon_\theta^{(t)}\left(\boldsymbol{x}_t\right)}_{\text {``direction pointing to } \boldsymbol{x}_t "}
\end{equation}
To show the correspondence between DDIM parameters and continuous-time SDE parameters, we follow one derivation in \cite{salimans2022progressive} by considering the ``predicted $x_0$'': note that define the predicted $x_0$ parameterization as: 
$$
\hat{x}_\theta\left(t,x\right) = \frac{x-\sqrt{1-\bar{\alpha}_t} \epsilon_\theta^{(t)}\left(x\right)}{\sqrt{\bar{\alpha}_t}},\text{ or , }\epsilon_\theta^{(t)}\left(x\right) = \frac{x- \sqrt{\bar{\alpha}_t}\hat{x}_\theta\left(t,x\right)}{\sqrt{1-\bar{\alpha}_t}},
$$
above \eqref{eq: DDIM update rule} can be rewritten as:
\begin{equation}
\label{DDIM rewritten}
\boldsymbol{x}_{t-1}=\frac{\sqrt{1-\bar{\alpha}_{t-1}}}{\sqrt{1-\bar{\alpha}_t}} \left(\boldsymbol{x}_t-\sqrt{\bar{\alpha}_t} \hat{x}_\theta\left(t,x\right)\right)+\sqrt{\bar{\alpha}_{t-1}} \cdot \hat{x}_\theta\left(t,x\right)
\end{equation}
Using parameterization $\sigma_t=\sqrt{1-\bar{\alpha}_{t}}$ and $\alpha_t = \sqrt{\bar{\alpha}_{t}}$, we have for $t-1 = s<t$:
\begin{equation}
\label{DDIM update rule continous}
X_s=\frac{\sigma_s}{\sigma_t}\left[X_t-\alpha_t \hat{x}_\theta\left(t,X_t\right)\right]+\alpha_s \hat{x}_\theta\left(t,X_t\right),
\end{equation}
which is the same as derived in \cite{kingma2021variationalDM,salimans2022progressive}.
\vskip 5pt
\subsubsection{ODE explanation by analyzing the derivative}
We further assume a VP diffusion process with $\alpha_t^2=1-\sigma_t^2=\operatorname{sigmoid}\left(\lambda_t\right)$ for $\lambda_t=\log \left[\alpha_t^2 / \sigma_t^2\right]$, in which $\lambda_t$ is known as the signal-to-noise ratio. Taking the derivative of \eqref{DDIM update rule continous} with respect to $\lambda_s$, assuming again a variance preserving diffusion process, and using $\frac{d \alpha_\lambda}{d \lambda}=\frac{1}{2} \alpha_\lambda \sigma_\lambda^2$ and $\frac{d \sigma_\lambda}{d \lambda}=-\frac{1}{2} \sigma_\lambda \alpha_\lambda^2$, gives
$$
\begin{aligned}
\frac{X_{\lambda_s}}{d \lambda_s} & =\frac{d \sigma_{\lambda_s}}{d \lambda_s} \frac{1}{\sigma_t}\left[X_t-\alpha_t \hat{x}_\theta\left(t,X_t\right)\right]+\frac{d \alpha_{\lambda_s}}{d \lambda_s} \hat{x}_\theta\left(t,X_t\right) \\
& =-\frac{1}{2} \alpha_s^2 \frac{\sigma_s}{\sigma_t}\left[X_t-\alpha_t \hat{x}_\theta\left(t,X_t\right)\right]+\frac{1}{2} \alpha_s \sigma_s^2 \hat{x}_\theta\left(t,X_t\right) .
\end{aligned}
$$

Evaluating this derivative at $s=t$ then gives
\begin{equation}
\label{DDIM derivative}
\begin{aligned}
\left.\frac{X_{\lambda_s}}{d \lambda_s}\right|_{s=t} & =-\frac{1}{2} \alpha_\lambda^2\left[X_\lambda-\alpha_\lambda \hat{x}_\theta\left(t,X_\lambda\right)\right]+\frac{1}{2} \alpha_\lambda \sigma_\lambda^2 \hat{x}_\theta\left(t,X_\lambda\right) \\
& =-\frac{1}{2} \alpha_\lambda^2\left[X_\lambda-\alpha_\lambda \hat{x}_\theta\left(t,X_\lambda\right)\right]+\frac{1}{2} \alpha_\lambda\left(1-\alpha_\lambda^2\right) \hat{x}_\theta\left(t,X_\lambda\right) \\
& =\frac{1}{2}\left[\alpha_\lambda \hat{x}_\theta\left(t,X_\lambda\right)-\alpha_\lambda^2 X_\lambda\right] .
\end{aligned}
\end{equation}
Recall that the forward process in terms of an SDE is defined as:
$$
\mathrm{d} X_t=f(t,X_t) \mathrm{d} t+g(t) \mathrm{d} B_t,\quad t\in [0,T]
$$
and \cite{Song20SGMbySDE} shows that backward of this diffusion process is an SDE, but shares the same marginal probability density of an associated probability flow ODE  (by taking $t:=T-t$) :
$$
\mathrm{d} X_t=\left[f(t,X_t)-\frac{1}{2} g^2(t) \nabla_x \log p(t,X_t)\right] \mathrm{d} t,\quad t\in [T,0]
$$
where in practice $\nabla_x \log p(t,x)$ is approximated by a learned denoising model using
\begin{equation}
\label{score_parameterization}
\nabla_x \log p(t,x) \approx s_{\theta}(t,x)=\frac{\alpha_t \hat{x}_\theta\left(t,x\right)-x}{\sigma_t^2} = -\frac{\epsilon_\theta^{(t)}\left(x\right)}{\sigma_t}.
\end{equation}
with two chosen noise scheduling parameters $\alpha_t$ and $\sigma_t$, and corresponding drift term $f(t,x)=\frac{d \log \alpha_t}{d t} x_t$ and diffusion term $g^2(t)=\frac{d \sigma_t^2}{d t}-2 \frac{d \log \alpha_t}{d t} \sigma_t^2$. 
\vskip 7pt
Further assuming a VP diffusion process with $\alpha_t^2=1-\sigma_t^2=\operatorname{sigmoid}\left(\lambda_t\right)$ for $\lambda_t=\log \left[\alpha_t^2 / \sigma_t^2\right]$, we get
$$
f(t,x)=\frac{d \log \alpha_t}{d t} x=\frac{1}{2} \frac{d \log \alpha_\lambda^2}{d \lambda} \frac{d \lambda}{d t} x=\frac{1}{2}\left(1-\alpha_t^2\right) \frac{d \lambda}{d t} x=\frac{1}{2} \sigma_t^2 \frac{d \lambda}{d t} x.
$$
Similarly, we get
$$
g^2(t)=\frac{d \sigma_t^2}{d t}-2 \frac{d \log \alpha_t}{d t} \sigma_t^2=\frac{d \sigma_\lambda^2}{d \lambda} \frac{d \lambda}{d t}-\sigma_t^4 \frac{d \lambda}{d t}=\left(\sigma_t^4-\sigma_t^2\right) \frac{d \lambda}{d t}-\sigma_t^4 \frac{d \lambda}{d t}=-\sigma_t^2 \frac{d \lambda}{d t}.
$$
Plugging these into the probability flow ODE then gives
\begin{equation}
\label{eq:DDIM ODE}
\begin{aligned}
\mathrm{d} X_t & =\left[f(t,X_t)-\frac{1}{2} g^2(t) \nabla_x \log p(t,x)\right] \mathrm{d} t \\
& =\frac{1}{2} \sigma_t^2\left[X_t+\nabla_x \log p(t,X_t)\right] \mathrm{d} \lambda_t .
\end{aligned}
\end{equation}
Plugging in our function approximation from Equation  \eqref{score_parameterization} gives
\begin{equation}
\label{eq:ODE_practical}
\begin{aligned}
\mathrm{d}  X_t & =\frac{1}{2} \sigma_t^2\left[X_t+\left(\frac{\alpha_t \hat{x}_\theta\left(t,X_t\right)-X_t}{\sigma_t^2}\right)\right] \mathrm{d} \lambda_t \\
& =\frac{1}{2}\left[\alpha_t \hat{x}_\theta\left(t,X_t\right)+\left(\sigma_t^2-1\right) X_t\right] \mathrm{d} \lambda_t \\
& =\frac{1}{2}\left[\alpha_t\hat{x}_\theta\left(t,X_t\right)-\alpha_t^2 X_t\right]\mathrm{d} \lambda_t .
\end{aligned}
\end{equation}
Comparison this with Equation \eqref{DDIM derivative} now shows that DDIM follows the probability flow ODE up to first order, and can thus be considered as an integration rule for this ODE. 

\subsubsection{Exponential Integrator Explanation}
In \cite{zhang2022fast} that the integration role above is referred as "exponential integrator" of \eqref{eq:ODE_practical}. We adopt two ways of derivations:
\vskip 5pt
(a) Notice that, if we treat the $\hat{x}_\theta\left(t,X_t\right)$ as a constant in \eqref{eq:ODE_practical} (or assume that it does not change w.r.p. $t$ along the ODE trajectory), we have:
\begin{equation}
\begin{aligned}
\mathrm{d}  X_t + \frac{1}{2}\alpha_t^2 X_t\mathrm{d} \lambda_t & =\hat{x}_\theta\left(t,X_t\right)\cdot\frac{1}{2} \alpha_t \mathrm{d} \lambda_t .
\end{aligned}
\end{equation}
Both sides multiplied by $1/\sigma_t$ and integrate from $t$ to $s$ yields:
\begin{equation}
\begin{aligned}
\frac{X_s}{\sigma_s}-\frac{X_t}{\sigma_t}=\hat{x}_\theta\left(t,X_t\right)\cdot\left(\exp(\frac{1}{2}\lambda_s)-\exp(\frac{1}{2}\lambda_t)\right)=\hat{x}_\theta\left(t,X_t\right)\cdot\left(\frac{\alpha_s}{\sigma_s}-\frac{\alpha_t}{\sigma_t}\right).
\end{aligned}
\end{equation}
which is thus
\begin{equation}
\begin{aligned}
X_s 
&=  \frac{\sigma_s}{\sigma_t}X_t+\left[\alpha_s-\alpha_t \frac{\sigma_s}{\sigma_t} \right] \hat{x}_\theta\left(t,X_t\right),
\end{aligned}
\end{equation}
which is the same as DDIM continuous-time intepretation as in \eqref{DDIM update rule continous}. 
\vskip 5 pt
(b) We also notice that we can also simplify the whole proof by treating the scaled score (same as in \cite{zhang2022fast})
\begin{equation}
\label{scaled_score}
\sigma_t\nabla_x \log p(t,x) \approx \sigma_t s_{\theta}(t,x)=\frac{\alpha_t \hat{x}_\theta\left(t,x\right)-x}{\sigma_t}
\end{equation}
as a constant in \eqref{eq:DDIM ODE} (or assume that it does not change w.r.p. $t$ along the ODE trajectory). Notice that from backward ODE, we have:
\begin{equation}
\begin{aligned}
\mathrm{d} X_t =\frac{1}{2} \sigma_t^2\left[X_t+\frac{1}{\sigma_t}\sigma_t\nabla_x \log p(t,X_t)\right] \mathrm{d} \lambda_t .
\end{aligned}
\end{equation}
Both sides multiplied by $1/\alpha_t$ and integrate from $t$ to $s$ yields:
\begin{equation}
\begin{aligned}
\frac{X_s}{\alpha_s}-\frac{X_t}{\alpha_t}=\left(\frac{\alpha_t \hat{x}_\theta\left(t,X_t\right)-X_t}{\sigma_t}\right)\cdot\left(-\frac{\sigma_s}{\alpha_s}+\frac{\sigma_t}{\alpha_t}\right).
\end{aligned}
\end{equation}
which is thus
\begin{equation}
\begin{aligned}
X_s 
&=  \frac{\sigma_s}{\sigma_t}X_t+\left[\alpha_s-\alpha_t \frac{\sigma_s}{\sigma_t} \right] \hat{x}_\theta\left(t,X_t\right),
\end{aligned}
\end{equation}
which is the same as DDIM continuous-time intepretation as in \eqref{DDIM update rule continous}. 
\vskip 7 pt
As a summary, treating the denoised mean or the noise predictor as the constants will both recovery the rule of DDIM.
\newpage

\section{Theorem Proofs}

\subsection{Proof of Theorem \ref{thm:Regularization as KL bound}}
\label{Proof of Regularization as KL bound}
The main proof technique relies on Girsanov's Theorem, which is similar to the argument in \cite{chen2022sampling}. First, we recall a consequence of Girsanov's theorem that can be obtained by combining Pages 136-139, Theorem 5.22, and Theorem 4.13 of Le Gall (2016).
\begin{thm} For $t \in[0, T]$, let $\mathcal{L}_t=\int_0^t b_s \mathrm{~d} B_s$ where $B$ is a $Q$-Brownian motion. Assume that $\mathbb{E}_Q \int_0^T\left\|b_s\right\|^2 \mathrm{~d} s<\infty$. Then, $\mathcal{L}$ is a $Q$-martingale in $L^2(Q)$. Moreover, if
\begin{equation}
\label{condition for Girsanov}
\mathbb{E}_Q \mathcal{E}(\mathcal{L})_T=1, \quad \text { where } \mathcal{E}(\mathcal{L})_t:=\exp \left(\int_0^t b_s \mathrm{~d} B_s-\frac{1}{2} \int_0^t\left\|b_s\right\|^2 \mathrm{~d} s\right),
\end{equation}
then $\mathcal{E}(\mathcal{L})$ is also a $Q$-martingale, and the process
\begin{equation}
t \mapsto B_t-\int_0^t b_s \mathrm{~d} s
\end{equation}
is a Brownian motion under $P:=\mathcal{E}(\mathcal{L})_T Q$, the probability distribution with density $\mathcal{E}(\mathcal{L})_T$ w.r.t. $Q$.
\end{thm}
If the assumptions of Girsanov's theorem are satisfied (i.e., the condition \eqref{condition for Girsanov}), we can apply Girsanov's theorem to $Q$ as the law of the following reverse process (we omit $c$ for brevity),
\begin{equation}
\label{eqn:ReverseSDEapprox by pretrain}
\mathrm{d} \overline{X}_t = \left(-f(T-t,\overline{X}_t) + \frac{1+\eta^2}{2}g^2(T-t) s_{\theta_{pre}}(T-t, \overline{X}_t)  \right) \mathrm{d}t + \eta g(T-t) \mathrm{d}B_t,\ \overline{X}_0\sim p_\infty(\cdot)
\end{equation}
and
\begin{equation}
b_t=\frac{1+\eta^2}{2\eta}g(T-t)\left[s_{\theta}(T-t, \overline{X}_t)-s_{\theta_{pre}}(T-t, \overline{X}_t)\right],
\end{equation}
where $t \in[0,T]$. This tells us that under $P=\mathcal{E}(\mathcal{L})_T Q$, there exists a Brownian motion $\left(\beta_t\right)_{t \in[0, T]}$ s.t.
\begin{equation}
\label{new BM under new measure}
\mathrm{d} B_t=\frac{1+\eta^2}{2\eta} g(T-t)\left[s_{\theta}(T-t, \overline{X}_t)-s_{\theta_{pre}}(T-t, \overline{X}_t)\right] \mathrm{d} t+\mathrm{d} \beta_t.
\end{equation}
Plugging \eqref{new BM under new measure} into \eqref{eqn:ReverseSDEapprox by pretrain} we have $P$-a.s.,
\begin{equation}
\mathrm{d} \overline{X}_t = \left(-f(T-t,\overline{X}_t) +\frac{1+\eta^2}{2\eta} g^2(T-t) s_{\theta}(T-t, \overline{X}_t)  \right) \mathrm{d}t + \eta g(T-t) \mathrm{d}\beta_t,\ \overline{X}_0\sim p_\infty(\cdot)
\end{equation}
In other words, under $P$, the distribution of $\overline{X}$ is the same as the distribution generated by current policy parameterized by $\theta$, i.e., $p_\theta(\cdot)=P_T=$ $\mathcal{E}(\mathcal{L})_T Q$. Therefore,
$$
\begin{aligned}
& D_{KL}\left(p_{\theta}\|p_{\theta_{pre}}\right)=\mathbb{E}_{P_T} \ln \frac{\mathrm{d} P_T}{\mathrm{d} Q_T}=\mathbb{E}_{P_T} \ln \mathcal{E}(\mathcal{L})_T \\
& =\mathbb{E}_{P_T}\left[\int_0^t b_s \mathrm{~d} B_s-\frac{1}{2} \int_0^t\left\|b_s\right\|^2\right] \\
& =\mathbb{E}_{P_T}\left[\int_0^t b_s \mathrm{~d} \beta_s+\frac{1}{2} \int_0^t\left\|b_s\right\|^2\right]\\
& =\frac{1}{2} \left(\frac{1+\eta^2}{2\eta}\right)^2\int_0^tg^2(T-t)\underbrace{\mathbb{E}_{P}\left\|s_{\theta}(T-t, \overline{X}_t)-s_{\theta_{pre}}(T-t, \overline{X}_t)\right\|^2}_{\epsilon_t^2}\mathrm{d}t\\
\end{aligned}
$$
Thus we can bound the discrepancy between distribution generated by the policy $\theta$ and the pretrained parameters $\theta_{pre}$ as 
\begin{equation}
D_{KL}(p_{\theta}\|p_{\theta_{pre}})\leq \frac{1}{2}\left(\frac{1+\eta^2}{2\eta}\right)^2\int_{0}^{T}g^2(T-t)\epsilon_t^2\mathrm{d}t
\end{equation}

\subsection{Proof of Theorem \ref{thm:PG formula}}
\label{Proof of PG formula}
First we include the policy gradient formula theorem for finite horizon in continuous time from \cite{jia2022policy_gradient}:
\begin{lem}[Theorem 5 of \cite{jia2022policy_gradient} when $R\equiv 0$]
\label{thm:Jia&Zhou PG}
Under some regularity conditions, given an admissible parameterized policy $\pi_{\theta}$, the policy gradient of the value function $V\left(t, x ; \pi^\theta\right)$ admits the following representation:
\begin{equation}
\label{eqn:Jia&Zhou PG}
\begin{aligned}
\frac{\partial}{\partial \theta} V(t, x ; \pi^\theta)= & \mathbb{E}^{\mathbb{P}}\left[\int _ { t } ^ { T } e ^ { - \beta ( s - t ) } \left\{\frac { \partial } { \partial \theta} \operatorname { l o g } \pi^ { \theta} ( a _ { s } ^ { \boldsymbol { \pi } ^ {\theta} } | s , X _ { s } ^ { \boldsymbol { \pi } ^ {\theta} } ) \left(\mathrm{d} V(s, X_s^{\pi^\theta} ; \pi^\theta)\right.\right.\right. \\
& \left.\left.\left.+\left[r_R(s, X_s^{\pi^\theta}, a_s^{\pi^\theta})-\beta V(s, X_s^{\pi^\theta} ; \pi^\theta)\right] \mathrm{d} s\right) \right\} \mid X_t^{\pi^\theta}=x\right], \quad(t, x) \in[0, T] \times \mathbb{R}^d
\end{aligned}
\end{equation}
in which we denote the regularized reward
$$
r_R(t, X_t^{\pi^\theta}, a_t^{\pi^\theta}) = \gamma(t) \|a_t^{\pi^\theta}-s^{\theta^{*}}(t,X_t)\|^2.
$$
\end{lem}
First, by applying It\^o's formula to $V(t,X_t)$, we have:
\begin{equation}
\mathrm{d}V(t,X_t) = \left[\frac{\partial V}{\partial t}(t,X_t)+\frac{1}{2}\sigma(t)^2\circ\frac{\partial^2 V}{\partial x^2}(t,X_t)\right]\mathrm{d}t+\frac{\partial V}{\partial x}(t,X_t)\mathrm{d}X_t.
\end{equation}
Further recall that:
\begin{equation}
q(t, x, a ; \pi)=\frac{\partial V}{\partial t}\left(t, x ; \pi\right)+\mathcal{H}\left(t, x, a, \frac{\partial V}{\partial x}\left(t,x ; \pi\right), \frac{\partial^2 V}{\partial x^2}\left(t,x ; \pi\right)\right)-\beta V\left(t,x ; \pi\right),
\end{equation}
this implies that (similar discussion also appeared in \cite{jia2022q_learning})
\begin{equation}
q\left(t, X_t^{\pi}, a_t^{\pi} ; \pi\right) \mathrm{d} t=\mathrm{d} J\left(t, X_t^{\pi} ; \pi\right)+r\left(t, X_t^{\pi}, a_t^{\pi}\right) \mathrm{d} t-\beta J\left(t, X_t^{\pi} ; \pi\right) \mathrm{d} t+\{\cdots\} \mathrm{d} B_t.
\end{equation}
Plug this equality back in \eqref{eqn:Jia&Zhou PG} yields:
\begin{equation}
\begin{aligned}
\frac{\partial}{\partial \theta} V(t, x ; \pi^\theta)= & \mathbb{E}^{\mathbb{P}}\left[\int _ { t } ^ { T } e ^ { - \beta ( s - t ) } \frac { \partial } { \partial \theta} \log \pi^ { \theta} ( a _ { s } ^ { \pi ^ {\theta} } | s , X _ { s } ^ {\pi^ {\theta} } ) q\left(t, X_t^{\pi}, a_t^{\pi} ; \pi\right)\mathrm{d} s \mid X_t^{\pi^\theta}=x\right],
\end{aligned}
\end{equation}
Let $t=0$, $\beta = -\alpha$ and further taking expectation to the initial distribution yields Theorem \ref{thm:PG formula}.



\end{appendices}
\end{document}